\pdfoutput=1

\documentclass[11pt,a4paper]{article}

\usepackage{EMNLP2023}

\usepackage{times}
\usepackage{latexsym}

\usepackage[T1]{fontenc}

\usepackage[utf8]{inputenc}

\usepackage{microtype}

\usepackage{inconsolata}

\usepackage{color}
\usepackage{algorithm}
\usepackage{algorithmic}
\usepackage{amsmath}
\usepackage{amsthm}
\usepackage{enumitem}
\usepackage{bm}
\usepackage{subfigure} 
\usepackage{nicefrac}
\usepackage{tabularx}
\usepackage{url}
\usepackage{booktabs}
\usepackage{multirow}
\usepackage{graphicx}
\usepackage{amssymb}
\usepackage{bbding}

\graphicspath{ {./images/} }

\newtheorem{prop}{Proposition}

\newcolumntype{L}[1]{>{\raggedright\let\newline\\\arraybackslash\hspace{0pt}}m{#1}}
\newcolumntype{C}[1]{>{\centering\let\newline  \\\arraybackslash\hspace{0pt}}m{#1}}
\newcolumntype{R}[1]{>{\raggedleft\let\newline \\\arraybackslash\hspace{0pt}}m{#1}}

\title{Relation-aware Ensemble Learning for Knowledge Graph Embedding}

\author{Ling Yue$^1$ \quad  Yongqi Zhang$^2$ \quad Quanming Yao$^1$ \quad Yong Li$^1$ \\
	\quad {\bf Xian Wu}$^3$ \quad {\bf Ziheng Zhang}$^3$ \quad {\bf Zhenxi Lin}$^3$ \quad {\bf Yefeng Zheng}$^3$ \\
	$^1$Department of Electronic Engineering, Tsinghua University, Beijing, China \\
	$^2$4Paradigm Inc., Beijing, China \\
	$^3$Jarvis Research Center, Tencent YouTu Lab, Shenzhen, China \\
	\texttt{\small {lingyue/qyaoaa/liyong07}@tsinghua.edu.cn,}
	\texttt{\small yzhangee@connect.ust.hk,} \\
	\texttt{\small {kevinxwu/zihengzhang/chalerislin/yefengzheng}@tencent.com}
}

\begin{document}
  \maketitle
  
  \begin{abstract}
    Knowledge graph (KG) embedding is a fundamental task
    in natural language processing,
    and various methods have been proposed to
    explore semantic patterns in distinctive ways.
    In this paper,
    we propose to learn an ensemble 
    by leveraging existing methods in a relation-aware manner.
    However,
    exploring these semantics using relation-aware ensemble leads to
    a much larger search space than general ensemble methods.
    To address this issue, we propose a divide-search-combine algorithm RelEns-DSC
    that searches the relation-wise ensemble weights independently.
    This algorithm has the same computation cost as general ensemble methods but with much better performance.
    Experimental results on benchmark datasets demonstrate 
    the effectiveness of the proposed method in efficiently searching relation-aware ensemble weights 
    and achieving state-of-the-art embedding performance.
    The code is public at \url{https://github.com/LARS-research/RelEns}.
    \footnote{L. Yue and Y. Zhang made equal contributions,
    Correspondence is to Q.Yao.}
  \end{abstract}

  \section{Introduction}
  
  Knowledge graph (KG) embedding is a popular method
  for inferring latent features
  and making predictions in incomplete KGs \cite{ji2021survey}.
  This  technique involves transforming entities and relations
  into low-dimensional vectors
  and using a scoring function \cite{bordes2013translating,wang2017knowledge}
  to assess the plausibility of a triplet
  (consisting of a head entity, a relation, and a tail entity).
  Well-known scoring functions,
  such as TransE~\cite{bordes2013translating},
  ComplEx~\cite{trouillon2017knowledge},
  ConvE~\cite{dettmers2018convolutional},
  and CompGCN~\cite{vashishth2019composition},
  have demonstrated remarkable success in learning from KGs.

  Ensemble learning
  is a well-known technique that improves the performance of machine learning tasks
  by combining and reweighting the predictions of multiple models~\cite{breiman1996bagging, wolpert1992stacked, dietterich2000ensemble}.
  Its effectiveness has also been verified in KG embedding
  by previous studies~\cite{krompass2015ensemble,wang2022probabilistic,rivas2022ensembles}.

  While designing different scoring functions to model various relation properties~\cite{ji2021survey,sun2019rotate,li2022house},
  such as symmetry, inversion, composition and hierarchy,
  is crucial for achieving good performance,
  existing ensemble methods do not reflect the relation-wise
  characteristics of different models.
  This
  motivates
  us to propose specific ensemble weights for different relations,
  named as RelEns problem, 
  in this paper.
  By doing so,
  different KG embedding models can specialize in different relations, 
  leading to improved performance.
  However,
  the number of parameters to be searched will linearly increase,
  which can significantly complicate the ensemble construction process
  especially for KGs with many relations.
  To alleviate the difficulty of searching for relation-wise ensemble weights,
  we propose DSC, an algorithm that
  \textbf{D}ivide the overall ensemble objective  into multiple sub-problems,
  \textbf{S}earch for the ensemble weights for each relation independently,
  and then \textbf{C}ombine the results.
  This approach significantly reduces the size of the search space
  and evaluation cost for individual sub-problems, 
  compared to the overall objective.
  
  In summary,
  we propose  RelEns-DSC, 
  a novel relation-aware ensemble learning method for KG embedding 
  that searches different ensemble weights independently for different relations, 
  using a divide-concur strategy.
  Empirically,
  RelEns-DSC significantly improves the performance on three benchmark datasets
  (WN18RR, FB15k-237, NELL-995)
  and achieves the first place on the
  large-scale leaderboards ogbl-biokg and ogbl-wikikg2.
  Our approach is more effective than general ensemble techniques,
  and it is more efficient with the divide-concur strategy under parallel computing.

  \section{Proposed Method}
  \label{sec:method}
  Denote a KG as $\mathcal G=(\mathcal V, \mathcal R, \mathcal D)$,
  where $\mathcal V$ contains $V$ entities (nodes),
  $\mathcal R$ contains $R$ types of relations between entities,
  and $\mathcal D = \{(h, r, t): h,t\in\mathcal V,  r\in\mathcal R\}$ contains the triplets (edges).
  $\mathcal D$ is split into three disjoint sets
  $\mathcal D_\text{tra}, \mathcal D_\text{val}, \mathcal D_\text{tst}$
  for training, validation and testing, respectively.
  
  The learning objective of a KG embedding model
  is to rank positive triplets higher than negative triplets,
  in order to accurately identify the potential positive triplets missed in the current graph~\cite{wang2017knowledge,ji2021survey}.
  
  Specifically, formulated as a tail prediction problem\footnote{Head prediction is conducted in the same way
  	with negative entities
  	$\mathcal N_h=\{e\in\mathcal V: (e,r,t)\notin \mathcal D\}$.
  	For simplicity,
  	we only use tail prediction as an example to introduce our method.}, the KG embedding model aims to rank the tail entity $t$ of a given triplet $x=(h,r,t)$, 
  which belongs to either $\mathcal D_\text{val}$ or $\mathcal D_\text{tst}$, 
  higher than a set of negative entities.
  The set of negative entities is defined as 
  $\mathcal N_t=\{e\in\mathcal V: (h,r,e)\notin \mathcal D\}$.
  The model $F(x)$ computes a score vector $\bm s$ for each entity 
  $e\in\{t\}\cup\mathcal N_t$,
  which indicates the degree of plausibility that the triplet $(h,r,e)$ is true.
  
  A ranking function
  $\Gamma\big(\bm s\big)$  
  is used to convert the scores $\bm s$ into
  a ranking list $\bm p=(p_1, \dots, p_C)$
  for the $C\!=\!1\!+\!|\mathcal N_t|$ entities.
  A smaller rank value implies the higher prediction priority.
  Following~\cite{bordes2013translating,trouillon2017knowledge,sun2019rotate,vashishth2019composition},
  we adopt mean reciprocal ranking (MRR) as the evaluation metric.
  Larger MRR indicates better performance.

  \subsection{Relation-wise Ensemble Problem}
  \label{ssec:relation-wise}
  
  We observe that
  embedding models may exhibit varying strengths in
  modeling different types of relations (see Appendix~\ref{app:relationproperty} for details).
  To account for this,
  we propose a novel approach that learns distinct weights for each relation, 
  based on the performance of models on validation set $\mathcal D_\text{val}$.
  Specifically,
  given $N$ trained KG embedding models,
  i.e., 
  $F_1, F_2, \dots, F_N$,
  and a set  of relations $\mathcal R$.
  We introduce a weight $\alpha_i^r\!\geq\! 0$ assigned to model $F_i$ for relation $r$ 
  and $M(\bm p, x) = 1/p_t$ for the reciprocal ranking of a given data point $x=(h,r,t)$.
  Let $\mathcal D^r_\text{val}$ denote as the subset of validation triplets whose relations are $r$.
  The objective of relation-wise ensemble can be written as follows:
  \begin{align}
     &
    \!\!\!\!\!
    \max_{ \{ \alpha_i^r  \}_{i=1,\dots,N}^{r = 1,\dots,R}}
    \sum\nolimits_{r\in\mathcal R}\sum\nolimits_{x_j \in \mathcal D_\text{val}^r}
    M\big(\Gamma(\bm p_j^r), x_j^r\big),              \label{eq:no-split} \\
     & \text{s.t.}\;
    \bm p_j^r = - \sum\nolimits_{i = 1}^N \alpha_i^r \Gamma\big( F_i( x_j^r )\big),
    \alpha_i^r \ge 0.
    \nonumber
  \end{align}
  For each triplet $x_j^r$ with relation $r$,
  we apply the ensemble weights $\alpha_i^r$ 
  to the ranking list $\Gamma(F_i(x_j^r))$ generated by the $i$-th model.
  The scales of scores vary significantly. Optimizing scores directly may be more challenging. 
  Additionally, since ranks have similar scales,
  the searched weights can better indicate the importance of the corresponding base model.
  Specifically,
  we obtain the ensembled score $\bm p_j^r = - \sum_{i = 1}^N \alpha_i^r \Gamma\big( F_i( x_j^r )\big)$,
  where ``$-$'' turns the ranks to scores, 
  indicating higher prediction priority with a higher value in $\bm p_j^r$.

  In particular,
  if the ensemble weights assigned for each model $F_i$ for all relations are identical,
  i.e.,
  $\alpha_i^1\!=\!\alpha_i^2\!=\!\cdots \!=\! \alpha_i^{R}$ for $i\!=\!1,\dots,N$,
  the objective in equation \eqref{eq:no-split} (denoted as \textit{RelEns-Basic})
  reduces to the general ensemble method (denoted as \textit{SimpleEns}).
  By optimizing the values of ${\alpha_i^r}$, the goal is to achieve higher MRR 
  performance on the validation set 
  $\mathcal D_\text{val}= \sum_{r\in\mathcal R} \mathcal D_\text{val}^r$.

  \subsection{Divide Search and Combine}
  
  Comparing with \textit{SimpleEns},
  \textit{RelEns-Basic}
  requires searching for $NR$ parameters.
  As MRR is a non-differential metric,
  zero-order optimization techniques,
  like random search and Bayesian optimization \cite{bergstra2011algorithms},
  are often used to solve Eq. \eqref{eq:no-split}.
  However,
  these algorithms usually involve sampling
  candidates in the search space,
  the complexity of which can grow exponentially with the
  search dimension due to the curse of dimensionality~\cite{koppen2000curse}.
  As a result,
  optimizing
  Eq. \eqref{eq:no-split} can be challenging.
  To address this issue,
  we propose Proposition~\ref{prop:sepa}, 
  which enables the separation of 
  the big problem Eq. \eqref{eq:no-split}
  into $R$ independent sub-problems.
  In the  divided problem $r$,
  there are only $N$ parameters $\{\alpha^r_i\}_{i=1,\dots,N}$
  to be searched.

  \begin{prop}[separable optimization problem]
    \label{prop:sepa}
    The optimal values of $\{ \alpha_i^r  \}_{i=1,\dots,N}^{r = 1,\dots,R}$ that are searched on $\mathcal D_\emph{val}$ in \eqref{eq:no-split}
    can be equated to the values of $\{\alpha^r_i\}_{i=1,\dots,N}$ 
    that are independently optimized on
    $\mathcal D_\emph{val}^r$ for each $r\in\mathcal R$
    via the following problem
    \begin{align}
       & \max_{\{\alpha^r_i\}_{i=1,\dots,N}}
      \sum\nolimits_{x_j \in \mathcal D_\emph{val}^r} M(\Gamma(\bm p_j^r), x_j^r),
      \label{eq:split}
      \\
       & \text{s.t.}\;
      \bm p_j^r =
      - \sum\nolimits_{i = 1}^N \alpha_i^r\Gamma\big(F_i( x_j^r )\big), \alpha_i^r \ge 0.
      \nonumber
    \end{align}
  \end{prop}

  The complete divide-search-and-combine procedures 
  are outlined in Algorithm~\ref{alg:relens}.
  By separably searching the divided problems,
  we can determine the optimal values of 
  $\{\alpha^r_i\}_{i=1,\dots,N}$ for each $r$ on the validation data $\mathcal D_\text{val}^r$.
  
  Finally, we combine the searched values of {$\{\alpha^r_i\}_{i=1,\dots,N}^{r=1,\dots,R}$}
  to compute the scores
  $\bm p_j^r =
    - \sum\nolimits_{i = 1}^N \alpha_i^r\Gamma\big(F_i( x_j^r )\big)$
  for $x_j^r\in\mathcal D_\text{tst}$ in order to evaluate the performance.

  \begin{algorithm}[ht]
    \caption{RelEns-DSC: Divide-search-combine algorithm for relation-wise ensemble.}
    \label{alg:relens}
    \begin{algorithmic}[1]
      \REQUIRE
      Base models $F_1, \dots, F_N$,
      ensemble parameters $\{\alpha^r_i, i=1,\dots,N, r\in\mathcal R\}$,
      dataset $\mathcal D_\text{val}$,
      ranking functions $\Gamma(\cdot)$ and $M(\cdot, \cdot)$.
      \STATE \textbf{Divide}:
      divide  $\mathcal D_\text{val}$ and $\{\alpha^r_i, i\!=\!1,\dots,N, r\!\in\!\mathcal R\}$ into
      $\big\{\mathcal D_\text{val}^r, \{\alpha^r_i, i\!=\!1,\dots,N\},$$r\!\in\!\mathcal R\big\}$.
        \STATE \textbf{Search}:
        \FOR{$r$ in $\mathcal R$ (can work in parallel)}
        \STATE search the values of $\{\alpha^r_i\}_{i=1,\dots,N}$ by solving  \eqref{eq:split};
        \ENDFOR
        \STATE  \textbf{Combine}:
        Combine the optimal values of $\{ \alpha_i^r  \}_{i=1,\dots,N}^{r = 1,\dots,R}$
        for evaluation;
        \RETURN MRR and Hit@$k$ on testing data $\mathcal D_\text{tst}$.
    \end{algorithmic}
  \end{algorithm}

  \subsection{Complexity Analysis}
  Assuming that the evaluation cost of $\Gamma(\cdot)$ and $M(\Gamma(\cdot), x)$ on a single data sample $x$ is a constant,
  the time complexity of ensemble learning is determined by two factors:
  (i) the number of data samples to be evaluated;
  (ii) the number of ensemble parameters to be sampled.
  For \textit{SimpleEns}, the complexity is $O(|\mathcal D_\text{val}|e^N)$.
  On the other hand,
  \textit{RelEns-Basic} in Eq. \eqref{eq:no-split} requires $O(|\mathcal D_\text{val}|e^{RN})$
  since the sampling complexity increases exponentially with the search dimension.
  In comparison,
  the complexity of \textit{RelEns-DSC} in Algorithm~\ref{alg:relens} is $O(|\mathcal D_\text{val}|e^N)$,
  which is on par with \textit{SimpleEns}.

  \section{Experiments}
  The experiments were implemented using Python and run on a 24GB NVIDIA GTX3090 GPU.
  
  As the ranking function $\Gamma(\cdot)$ and MRR are non-differentiable,
  We chose the widely used Bayesian optimization technique, 
  Tree-Parzen Estimator (TPE)~\cite{bergstra2015hyperopt}, 
  to solve the maximization problems in Eq. \eqref{eq:no-split} and Eq. \eqref{eq:split},
  the details of which are provided in the Appendix~\ref{app:tpe}.
  
  \subsection{Datasets}
  
  We conduct experiments on commonly studied datasets for KG,
  including WN18RR~\cite{dettmers2018convolutional}, FB15k-237~\cite{toutanova2015observed}, and NELL-995~\cite{xiong2017deeppath}.
  Additionally, we apply the  \textit{RelEns-DSC} on OGB~\cite{hu2020open} datasets ogbl-biokg and ogbl-wikikg2.
  Details of statistics are in Appendix~\ref{app:datasets}.
  
  \subsection{Experimental Setup}
  \paragraph{Base models.}
  We select some representative embedding models as our base models
  $F_i$,
  including:
  (i) translational distance models TransE \cite{bordes2013translating}, 
  RotatE \cite{sun2019rotate}, HousE \cite{li2022house};
  (ii) bilinear model ComplEx \cite{trouillon2017knowledge};
  (iii) neural network model ConvE \cite{dettmers2018convolutional};
  and (iv) GNN based model CompGCN \cite{vashishth2019composition}.
  For the OGB datasets,
  we select the top-3 methods from the OGB leaderboard\footnote{\url{https://ogb.stanford.edu/docs/leader_linkprop/}.}
  up to October 1st in 2023.
%

  \paragraph{Evaluation Metric.}
  Four  evaluation metrics (MRR and Hit@\{1,3,10\}) are reported
  for the benchmarks WN18RR, FB15k-237 and NELL-995.
  For OGB datasets ogbl-biokg and ogbl-wikikg2,
  we report MRR results to keep consistent with the leaderboard.

  \paragraph{Hyperparameters.}
  To compare on the general benchmarks, 
  we use the fine-tuned hyperparameters reported by KGTuner~\cite{zhang2022kgtuner}.
  For top three methods on OGB leaderboard, 
  we use their reported hyperparameters.
  Details of these settings are in Appendix~\ref{app:hp}.

    \begin{table*}[ht]
  	\centering
  	\caption{Performance comparison on WN18RR, FB15k-237 and NELL-995 datasets.}
  	\small
  	\setlength\tabcolsep{3.2pt}
  	\renewcommand{\arraystretch}{1.2}
  	\vspace{-8px}
  	\begin{tabular}{c|C{26pt}C{26pt}C{26pt}C{30pt}|C{26pt}C{26pt}C{26pt}C{30pt}|C{26pt}C{26pt}C{26pt}C{30pt}}
  		\toprule
  		Dataset      & \multicolumn{4}{c|}{WN18RR} & \multicolumn{4}{c|}{FB15k-237} & \multicolumn{4}{c}{NELL-995}                                                                                                                                                                   \\ \midrule
  		Model        & MRR                         & Hit@1                          & Hit@3                        & Hit@10          & MRR             & Hit@1           & Hit@3           & Hit@10          & MRR             & Hit@1           & Hit@3           & Hit@10          \\ \midrule
  		TransE       & 0.2337                      & 0.0329                         & 0.3993                       & 0.5440          & 0.3277          & 0.2284          & 0.3687          & 0.5229          & 0.5072          & 0.4242          & 0.5593          & 0.6482          \\
  		RotatE       & 0.4772                      & 0.4236                         & 0.4982                       & 0.5799          & 0.3406          & 0.2468          & 0.3746          & 0.5284          & 0.5260          & 0.4658          & 0.5605          & 0.6260          \\
  		HousE      & 0.5103                      & 0.4644                         & 0.5258                       & 0.6023          & 0.3612          & 0.2658         & 0.3991          & 0.5504          & 0.5193          & 0.4581          & 0.5559          & 0.6178          \\
  		ComplEx      & 0.4833                      & 0.4403                         & 0.5029                       & 0.5613          & 0.3506          & 0.2606          & 0.3877          & 0.5283          & 0.5069          & 0.4423          & 0.5406          & 0.6107          \\
  		ConvE        & 0.4370                      & 0.3993                         & 0.4483                       & 0.5163          & 0.3333          & 0.2404          & 0.3667          & 0.5227          & 0.5294          & 0.4517          & 0.5782          & 0.6595          \\
  		CompGCN      & 0.4609                      & 0.4285                         & 0.4698                       & 0.5265          & 0.3355          & 0.2435          & 0.3715          & 0.5157          & 0.5167          & 0.4493          & 0.5617          & 0.6286          \\ 
  		\midrule
  		SimpleEns    & 0.5121   {\scriptsize$\pm$0.0007}                   & 0.4670    {\scriptsize$\pm$0.0004}                     & 0.5289   {\scriptsize$\pm$0.0015}                     & 0.6021   {\scriptsize$\pm$0.0017}        & 0.3621   {\scriptsize$\pm$.0.0011}        & 0.2683    {\scriptsize$\pm$0.0013}        & 0.3977  {\scriptsize$\pm$0.0018}         & 0.5525 {\scriptsize$\pm$0.0012}         & 0.5416  {\scriptsize$\pm$0.0032}         & 0.4758  {\scriptsize$\pm$0.0005}         & 0.5823 {\scriptsize$\pm$0.0026}          & 0.6601  {\scriptsize$\pm$0.0048}         \\ \midrule
  		RelEns-DSC & \textbf{0.5201}   {\scriptsize$\pm$0.0005}          & \textbf{0.4770} {\scriptsize$\pm$0.0003}                & \textbf{0.5375} {\scriptsize$\pm$0.0009}              & \textbf{0.6039} {\scriptsize$\pm$0.0015} & \textbf{0.3680} {\scriptsize$\pm$0.0008} & \textbf{0.2746} {\scriptsize$\pm$0.0014} & \textbf{0.4046} {\scriptsize$\pm$0.0012} & \textbf{0.5554} {\scriptsize$\pm$0.0010} & \textbf{0.5499} {\scriptsize$\pm$0.0017} & \textbf{0.4823} {\scriptsize$\pm$0.0013} & \textbf{0.5901} {\scriptsize$\pm$0.0022} & \textbf{0.6609} {\scriptsize$\pm$0.0035} \\
  		Relative $\uparrow$ & 1.56\% & 2.14\%   & 1.63\% & 0.3\% & 1.63\% & 2.35\%  & 1.73\%  &  0.52\%  &  1.53\%  & 1.37\%  & 1.34\%   & 0.44\%  \\   \bottomrule
  	\end{tabular}
  	\label{tab:performance}
  \end{table*}

  \begin{table*}[ht]
	\centering
	\caption{Comparison of MRR performance on ogbl-biokg and ogbl-wikikg2 datasets.}
	\small
	\vspace{-8px}
	\setlength\tabcolsep{3pt}
	\begin{tabular}{c|C{70px}cc|C{90px}cc}
		\toprule
		Dataset & \multicolumn{3}{c|}{ogbl-biokg} & \multicolumn{3}{c}{ogbl-wikikg2}                                     \\ \midrule
		&  Model  name     & Valid                           & Test                          
		&  Model  name    & Valid           & Test            \\ \midrule
		Top1     &   {AutoBLM \cite{zhang2022bilinear}}              & {0.8548}                          & {0.8543}                   &  StarGraph+TripleRE +Text~\cite{yao2023ogb}        & {0.7439}          & 0.7302          \\ \midrule
		Top2     &  ComplEx-RP \cite{chen2021relation}                & 0.8497                          & 0.8494                  & InterHT+ \cite{wang2022interht}         & 0.7420          & {0.7309}          \\  \midrule
		Top3         &   TripleRE~\cite{yu2022triplere}           & 0.8361                          & 0.8348                    &  StarGraph+TripleRE \cite{listargraph}        & 0.7291          & 0.7193          \\ \midrule
		SimpleEns        & --        & 0.9117{\scriptsize$\pm$0.0002}                         & 0.9112{\scriptsize$\pm$0.0003}            & --             & 0.7509{\scriptsize$\pm$0.0009}         & 0.7392{\scriptsize$\pm$0.0011}          \\  \midrule
		RelEns-DSC           & --          & \textbf{0.9627{\scriptsize$\pm$0.0004}}                 & \textbf{0.9618{\scriptsize$\pm$0.0002}}            & --        & \textbf{0.7541{\scriptsize$\pm$0.0007}} & \textbf{0.7430{\scriptsize$\pm$0.0010}} \\
		Relative
		$\uparrow$ & --  & 	5.59\%	&	5.55\%	& --  	&		0.43\%	&	0.51\%	\\
		\midrule
	\end{tabular}
	\label{tab:leaderboard}
\end{table*}


  \subsection{Performance Comparison}
  \label{ssec:performance}
  
  Table~\ref{tab:performance} and Table~\ref{tab:leaderboard} present 
  the testing performance comparison.
   \textit{SimpleEns} is the variant introduced in Section~\ref{ssec:relation-wise}. We observe that \textit{SimpleEns}
  consistently outperforms the base models
  by weighting different models according to their learning ability.
  The proposed method \textit{RelEns-DSC} surpasses \textit{SimpleEns} 
  by a large margin,
  verifying the effectiveness of
  considering relation-specific ensemble weights 
  for KG embedding.

The top models on ogbl-biokg are more diverse than ogbl-wikikg2.
On ogbl-biokg,
AutoBLM and ComplEx are bilinear models, while TripleRE is a translational model.
The training framework of the three models are also different.
In comparison,
the top three methods on ogbl-wikikg2
are all translational models with similar approaches of sharing entity embeddings.
As a result,
the variations of relation-wise performance of the top three models
on ogbl-biokg are larger than ogbl-wikikg2 
(with std 0.0452 vs. 0.0261).
This can explain why the relation-wise ensemble is more significant on ogbl-biokg than ogbl-wikikg2.
  
\begin{figure}[h]
    \centering
    \includegraphics[width=1.0\columnwidth]{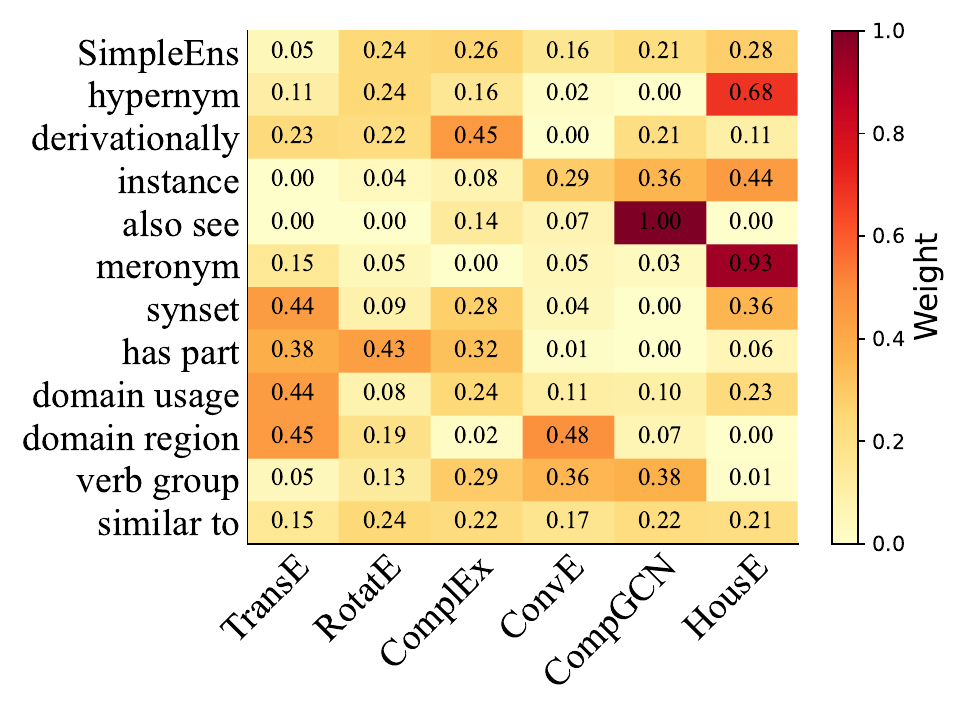}
	 \vspace{-14px}
    \caption{Ensemble weights of \textit{SimplEns} and \textit{RelEns-DSC} on the WN18RR dataset.}
  	\vspace{-5px}
    \label{fig:model_weights}
  \end{figure}

  Furthermore,
  we illustrate the ensemble weights of 
  \textit{SimpleEns} and \textit{RelEns-DSC} on the WN18RR dataset in Figure~\ref{fig:model_weights},
  which shows that \textit{RelEns-DSC} learns relation-specific ensemble weights,
  which contributes to its superior performance.

  \subsection{Efficiency Comparison}
  \label{sec:efficiency}
  We compare the learning curves (highest MRR yet searched vs. running time) of \textit{SimpleEns},
  \textit{RelEns-Basic}, and \textit{RelEns-DSC} on NELL-995 in Figure~\ref{fig:learning-curve} (the curves of other datasets are in Appendix~\ref{app:efficiency}).
  The ensemble weights for all three methods are initialized as $\nicefrac{1}{N}$.
  We denote the number of parameter configurations for TPE to search as $Q$,
  and show the results of  $Q=50$ and $Q=100$.
  
  \begin{figure*}[ht]
    \centering
    \includegraphics[height=0.52cm]{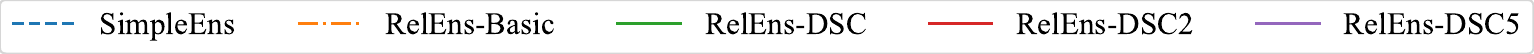}
    \subfigure[$Q= 50$]
    {
    	\includegraphics[height=5.47cm]{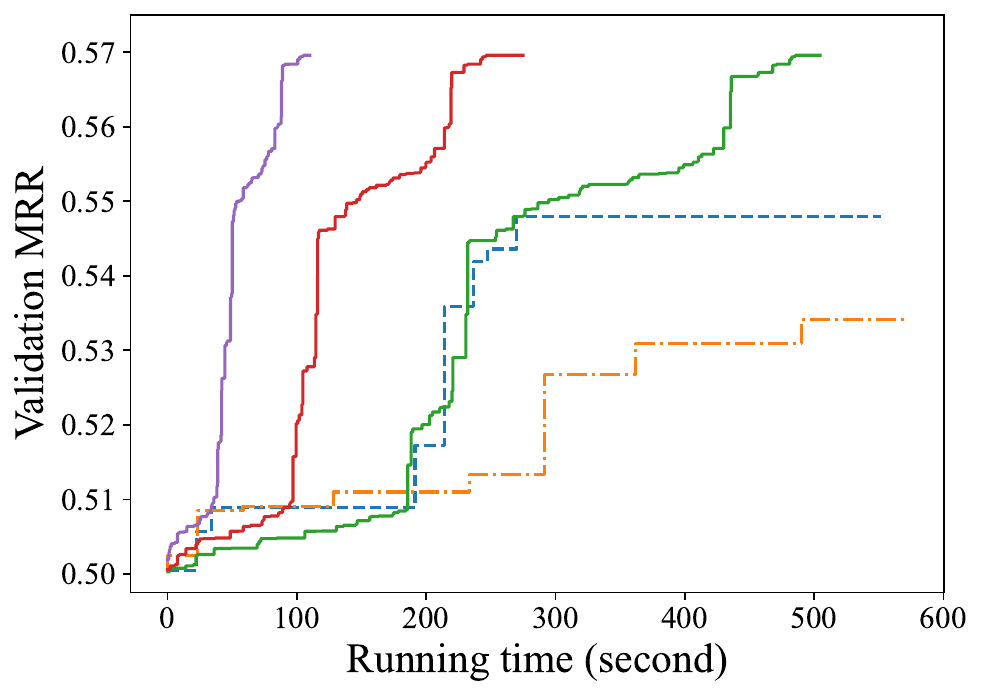}
      \label{fig:curve-50}}
    \hspace{-8pt}
    \subfigure[$Q= 100$]
    {\includegraphics[height=5.55cm]{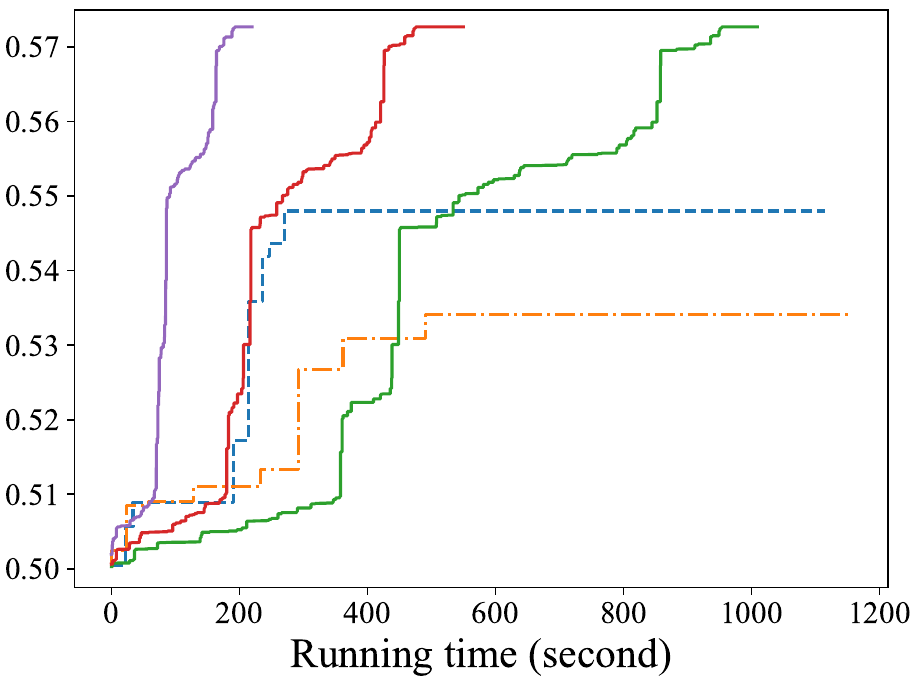}
      \label{fig:curve-100}}
  	\vspace{-10px}
    \caption{Learning curves of different ensemble methods on NELL-995.
       \textit{RelEns-DSC2/5} indicates the number of threads \textit{(2/5)} used for parallel computing.	
    }
    \label{fig:learning-curve}
    \vspace{-8px}
  \end{figure*}
  
  Based on the results,
  \textit{RelEns-Basic} is much worse than \textit{SimpleEns},
  since the search complexity of \textit{RelEns-Basic} increases exponentially.
  At the beginning of searching,
  \textit{RelEns-DSC} is inferior to both \textit{SimpleEns} and \textit{RelEns-Basic}
  since it only searches the weights on a few relations, while others are unchanged.
  Over time, 
  the performance of \textit{RelEns-DSC} has improved significantly 
  as more and more relations have found their optimal values.
  Increasing $Q$ from 50 to 100 did not lead to any improvement in the performance of \textit{SimpleEns} and \textit{RelEns-Basic}. 
  However, \textit{RelEns-DSC} was able to achieve better overall performance 
  since increasing $Q$ allows the sub-problems to be more sufficiently solved with more iterations.
  In addition,
  \textit{RelEns-DSC} can be benefited by parallel computing on the relation level,
  further improving efficiency.

  \subsection{Ablation Study}
  
    \begin{table}[ht]
  	\centering
  	\caption{Ablation study on WN18RR, FB15k-237 and NELL-995 dataset.
  		``H@10'' is short for Hit@10.}
  	\small
  	\renewcommand{\arraystretch}{1.1}
  	\vspace{-8px}
  	\setlength\tabcolsep{2.6pt}
  	\begin{tabular}{c|cc|cc|cc}
  		\toprule
  		Dataset    & \multicolumn{2}{c|}{WN18RR} & \multicolumn{2}{c|}{FB15k-237} & \multicolumn{2}{c}{NELL-995}                                                       \\ \midrule
  		Model      & MRR                         & H@10                          & MRR                          & H@10            & MRR             & H@10            \\ \midrule
  		Mean       & .4963                      & .5746                        & .3572                       & .5457          & .5412          & .6622          \\
  		Stacking   & .4952                      & .5751                        & .3563                       & .5433          & .5365          & .6669          \\
  		SimpleEns  & .5121                      & .6021                        & .3621                       & .5525          & .5416          & .6601          \\
  		MRR-Mean  & .5143                      & .6028                        & .3645                       & .5547          & .5460          & .6604          \\
  		RelEns-DSC & \textbf{.5201}   & \textbf{.6039}   & \textbf{.3680}    & \textbf{.5554} & \textbf{.5499} & \textbf{.6609} \\ \bottomrule
  	\end{tabular}
  	\label{tab:ensemble}
  \end{table}

  Table~\ref{tab:ensemble} shows the performance comparison of multiple variants of \textit{RelEns-DSC} on the three benchmark datasets.
  Due to space limit, results of Hit@\{1,3\}
  and the implementation details of the variants are provided in Appendix B.2.
  The stacking method (\textit{Stacking}), arithmetic mean method (\textit{Mean}) 
  and MRR-based weighted mean method (\textit{MRR-Mean})
  have poorer performance compared to \textit{RelEns-DSC}.
  This indicates the importance of searching for ensemble weights
  with TPE technique.
  \textit{Stacking} performs the worst since 
  the non-differentiable metric MRR cannot directly optimized.
  In particular,
  considering relation-specific ensemble weights,
  \textit{RelEns-DSC} 
  can lead to better performance than the general ensemble methods.

  \section{Conclusion}
  
  This paper introduces a novel ensemble method, Relation-aware Ensemble 
  with Divide-Search-Combine (RelEns-DSC) for KG embedding.
  The proposed RelEns-DSC learns relation-specific ensemble weights for different models
  and efficiently searches the weights using the divide-concur strategy.
  Empirical results demonstrate that our proposed method 
  outperforms existing ensemble methods for
  KG embedding,
  in both effectiveness and efficiency.
  
  
\paragraph{Limitations.}
  The proposed method mainly addresses the ensemble problem for entity prediction tasks in knowledge graph completion. 
  However, it does not effectively address 
  the other graph learning tasks,
  such as entity/node classification,
  relation prediction, 
  and graph classification.
  In addition,
  the significance of RelEns-DSC is under the case of
  multi-relational graphs like knowledge graph and heterogeneous graph,
  thus is not well adapted to homogeneous graph with single edge type.
  
 \section*{Acknowledgements}
 
 The work was performed when L. Yue was an research engineer in LARS group.
 Q. Yao was in part sponsored by NSFC (No. 92270106) and CCF-Tencent Open Research Fund.
  
 \clearpage
  
\bibliographystyle{acl_natbib}
\bibliography{anthology,custom}

\begin{thebibliography}{29}
\expandafter\ifx\csname natexlab\endcsname\relax\def\natexlab#1{#1}\fi

\bibitem[{Bergstra et~al.(2011)Bergstra, Bardenet, Bengio, and
  K{\'e}gl}]{bergstra2011algorithms}
James Bergstra, R{\'e}mi Bardenet, Yoshua Bengio, and Bal{\'a}zs K{\'e}gl.
  2011.
\newblock Algorithms for hyper-parameter optimization.
\newblock In \emph{NIPS}, pages 2546--2554.

\bibitem[{Bergstra et~al.(2015)Bergstra, Komer, Eliasmith, Yamins, and
  Cox}]{bergstra2015hyperopt}
James Bergstra, Brent Komer, Chris Eliasmith, Dan Yamins, and David~D Cox.
  2015.
\newblock Hyperopt: A python library for model selection and hyperparameter
  optimization.
\newblock \emph{Comput. Sci. Discov.}, 8(1):014008.

\bibitem[{Bordes et~al.(2013)Bordes, Usunier, Garcia-Duran, Weston, and
  Yakhnenko}]{bordes2013translating}
Antoine Bordes, Nicolas Usunier, Alberto Garcia-Duran, Jason Weston, and Oksana
  Yakhnenko. 2013.
\newblock Translating embeddings for modeling multi-relational data.
\newblock In \emph{NeurIPS}, pages 2787--2795.

\bibitem[{Breiman(1996)}]{breiman1996bagging}
Leo Breiman. 1996.
\newblock Bagging predictors.
\newblock \emph{Machine Learning}, 24:123--140.

\bibitem[{Chen et~al.(2021)Chen, Minervini, Riedel, and
  Stenetorp}]{chen2021relation}
Yihong Chen, Pasquale Minervini, Sebastian Riedel, and Pontus Stenetorp. 2021.
\newblock Relation prediction as an auxiliary training objective for improving
  multi-relational graph representations.
\newblock \emph{arXiv:2110.02834}.

\bibitem[{Dettmers et~al.(2018)Dettmers, Minervini, Stenetorp, and
  Riedel}]{dettmers2018convolutional}
Tim Dettmers, Pasquale Minervini, Pontus Stenetorp, and Sebastian Riedel. 2018.
\newblock Convolutional {2D} knowledge graph embeddings.
\newblock In \emph{AAAI}, volume~32.

\bibitem[{Dietterich(2000)}]{dietterich2000ensemble}
Thomas~G Dietterich. 2000.
\newblock Ensemble methods in machine learning.
\newblock In \emph{MCS}, pages 1--15. Springer.

\bibitem[{Hu et~al.(2020)Hu, Fey, Zitnik, Dong, Ren, Liu, Catasta, and
  Leskovec}]{hu2020open}
Weihua Hu, Matthias Fey, Marinka Zitnik, Yuxiao Dong, Hongyu Ren, Bowen Liu,
  Michele Catasta, and Jure Leskovec. 2020.
\newblock Open graph benchmark: Datasets for machine learning on graphs.
\newblock \emph{NeurIPS}.

\bibitem[{Ji et~al.(2021)Ji, Pan, Cambria, Marttinen, and
  Philip}]{ji2021survey}
Shaoxiong Ji, Shirui Pan, Erik Cambria, Pekka Marttinen, and S~Yu Philip. 2021.
\newblock A survey on knowledge graphs: Representation, acquisition, and
  applications.
\newblock \emph{IEEE transactions on neural networks and learning systems},
  33(2):494--514.

\bibitem[{K{\"o}ppen(2000)}]{koppen2000curse}
Mario K{\"o}ppen. 2000.
\newblock The curse of dimensionality.
\newblock In \emph{5th Online World Conference on Soft Computing in Industrial
  Applications}, volume~1, pages 4--8.

\bibitem[{Krompa{\ss} and Tresp(2015)}]{krompass2015ensemble}
Denis Krompa{\ss} and Volker Tresp. 2015.
\newblock Ensemble solutions for link-prediction in knowledge graphs.
\newblock In \emph{PKDD ECML 2nd Workshop on Linked Data for Knowledge
  Discovery}.

\bibitem[{Li et~al.()Li, Gao, Feng, Deng, and Yin}]{listargraph}
Hongzhu Li, Xiangrui Gao, Linhui Feng, Yafeng Deng, and Yuhui Yin.
\newblock Stargraph: Knowledge representation learning based on incomplete
  two-hop subgraph.
\newblock Technical report.

\bibitem[{Li et~al.(2022)Li, Zhao, Li, He, Wang, Liu, Sun, Wang, Deng, Shen
  et~al.}]{li2022house}
Rui Li, Jianan Zhao, Chaozhuo Li, Di~He, Yiqi Wang, Yuming Liu, Hao Sun,
  Senzhang Wang, Weiwei Deng, Yanming Shen, et~al. 2022.
\newblock House: Knowledge graph embedding with householder parameterization.
\newblock In \emph{ICML}, pages 13209--13224. PMLR.

\bibitem[{Pedregosa et~al.(2011)Pedregosa, Varoquaux, Gramfort, Michel,
  Thirion, Grisel, Blondel, Prettenhofer, Weiss, Dubourg
  et~al.}]{pedregosa2011scikit}
Fabian Pedregosa, Ga{\"e}l Varoquaux, Alexandre Gramfort, Vincent Michel,
  Bertrand Thirion, Olivier Grisel, Mathieu Blondel, Peter Prettenhofer, Ron
  Weiss, Vincent Dubourg, et~al. 2011.
\newblock Scikit-learn: Machine learning in python.
\newblock \emph{the Journal of machine Learning research}, 12:2825--2830.

\bibitem[{Rivas-Barragan et~al.(2022)Rivas-Barragan, Domingo-Fern{\'a}ndez,
  Gadiya, and Healey}]{rivas2022ensembles}
Daniel Rivas-Barragan, Daniel Domingo-Fern{\'a}ndez, Yojana Gadiya, and David
  Healey. 2022.
\newblock Ensembles of knowledge graph embedding models improve predictions for
  drug discovery.
\newblock \emph{Briefings in Bioinformatics}, 23(6):bbac481.

\bibitem[{Snoek et~al.(2012)Snoek, Larochelle, and Adams}]{snoek2012practical}
Jasper Snoek, Hugo Larochelle, and Ryan~P Adams. 2012.
\newblock Practical bayesian optimization of machine learning algorithms.
\newblock \emph{NIPS}, 25.

\bibitem[{Sun et~al.(2019)Sun, Deng, Nie, and Tang}]{sun2019rotate}
Zhiqing Sun, Zhi-Hong Deng, Jian-Yun Nie, and Jian Tang. 2019.
\newblock Rotate: Knowledge graph embedding by relational rotation in complex
  space.
\newblock In \emph{ICLR}.

\bibitem[{Toutanova and Chen(2015)}]{toutanova2015observed}
Kristina Toutanova and Danqi Chen. 2015.
\newblock Observed versus latent features for knowledge base and text
  inference.
\newblock In \emph{CVSC}, pages 57--66.

\bibitem[{Trouillon et~al.(2017)Trouillon, Dance, Welbl, Riedel, Gaussier, and
  Bouchard}]{trouillon2017knowledge}
Th{\'e}o Trouillon, Christopher~R Dance, Johannes Welbl, Sebastian Riedel,
  {\'E}ric Gaussier, and Guillaume Bouchard. 2017.
\newblock Knowledge graph completion via complex tensor factorization.
\newblock \emph{JMLR}, 18(1):4735--4772.

\bibitem[{Vashishth et~al.(2020)Vashishth, Sanyal, Nitin, and
  Talukdar}]{vashishth2019composition}
Shikhar Vashishth, Soumya Sanyal, Vikram Nitin, and Partha Talukdar. 2020.
\newblock Composition-based multi-relational graph convolutional networks.
\newblock \emph{ICLR}.

\bibitem[{Wang et~al.(2022{\natexlab{a}})Wang, Meng, Wang, Wu, Che, Wang, Chen,
  and Liu}]{wang2022interht}
Baoxin Wang, Qingye Meng, Ziyue Wang, Dayong Wu, Wanxiang Che, Shijin Wang,
  Zhigang Chen, and Cong Liu. 2022{\natexlab{a}}.
\newblock {InterHT}: Knowledge graph embeddings by interaction between head and
  tail entities.
\newblock \emph{arXiv:2202.04897}.

\bibitem[{Wang et~al.(2017)Wang, Mao, Wang, and Guo}]{wang2017knowledge}
Quan Wang, Zhendong Mao, Bin Wang, and Li~Guo. 2017.
\newblock Knowledge graph embedding: A survey of approaches and applications.
\newblock \emph{TKDE}, 29(12):2724--2743.

\bibitem[{Wang et~al.(2022{\natexlab{b}})Wang, Chen, Zhang, and
  Wang}]{wang2022probabilistic}
Yinquan Wang, Yao Chen, Zhe Zhang, and Tian Wang. 2022{\natexlab{b}}.
\newblock A probabilistic ensemble approach for knowledge graph embedding.
\newblock \emph{Neurocomputing}, 500:1041--1051.

\bibitem[{Wolpert(1992)}]{wolpert1992stacked}
David~H Wolpert. 1992.
\newblock Stacked generalization.
\newblock \emph{Neural Networks}, 5(2):241--259.

\bibitem[{Xiong et~al.(2017)Xiong, Hoang, and Wang}]{xiong2017deeppath}
Wenhan Xiong, Thien Hoang, and William~Yang Wang. 2017.
\newblock {DeepPath}: A reinforcement learning method for knowledge graph
  reasoning.
\newblock \emph{arXiv:1707.06690}.

\bibitem[{Yao et~al.(2023)Yao, Peng, Liu, Cai, Ji, He, and Cheng}]{yao2023ogb}
Liang Yao, Jiazhen Peng, Qiang Liu, Hongyun Cai, Shenggong Ji, Feng He, and
  Xu~Cheng. 2023.
\newblock Technical report for {OGB} link property prediction: ogbl-wikikg2.

\bibitem[{Yu et~al.(2022)Yu, Luo, Liu, Lin, Li, and Deng}]{yu2022triplere}
Long Yu, Zhicong Luo, Huanyong Liu, Deng Lin, Hongzhu Li, and Yafeng Deng.
  2022.
\newblock {TripleRE}: Knowledge graph embeddings via tripled relation vectors.
\newblock \emph{arXiv:2209.08271}.

\bibitem[{Zhang et~al.(2022{\natexlab{a}})Zhang, Yao, and
  Kwok}]{zhang2022bilinear}
Yongqi Zhang, Quanming Yao, and James~T Kwok. 2022{\natexlab{a}}.
\newblock Bilinear scoring function search for knowledge graph learning.
\newblock \emph{TPAMI}.

\bibitem[{Zhang et~al.(2022{\natexlab{b}})Zhang, Zhou, Yao, and
  Li}]{zhang2022kgtuner}
Yongqi Zhang, Zhanke Zhou, Quanming Yao, and Yong Li. 2022{\natexlab{b}}.
\newblock {KGTuner}: Efficient hyper-parameter search for knowledge graph
  learning.
\newblock In \emph{ACL}.

\end{thebibliography}
  
  \clearpage
  \appendix

  \onecolumn
  
  \section{Supplementary Materials for the Method}
  
  \subsection{Relation-wise Ensemble}
\begin{figure}[ht]
    \centering
    \includegraphics[width=0.8\columnwidth]{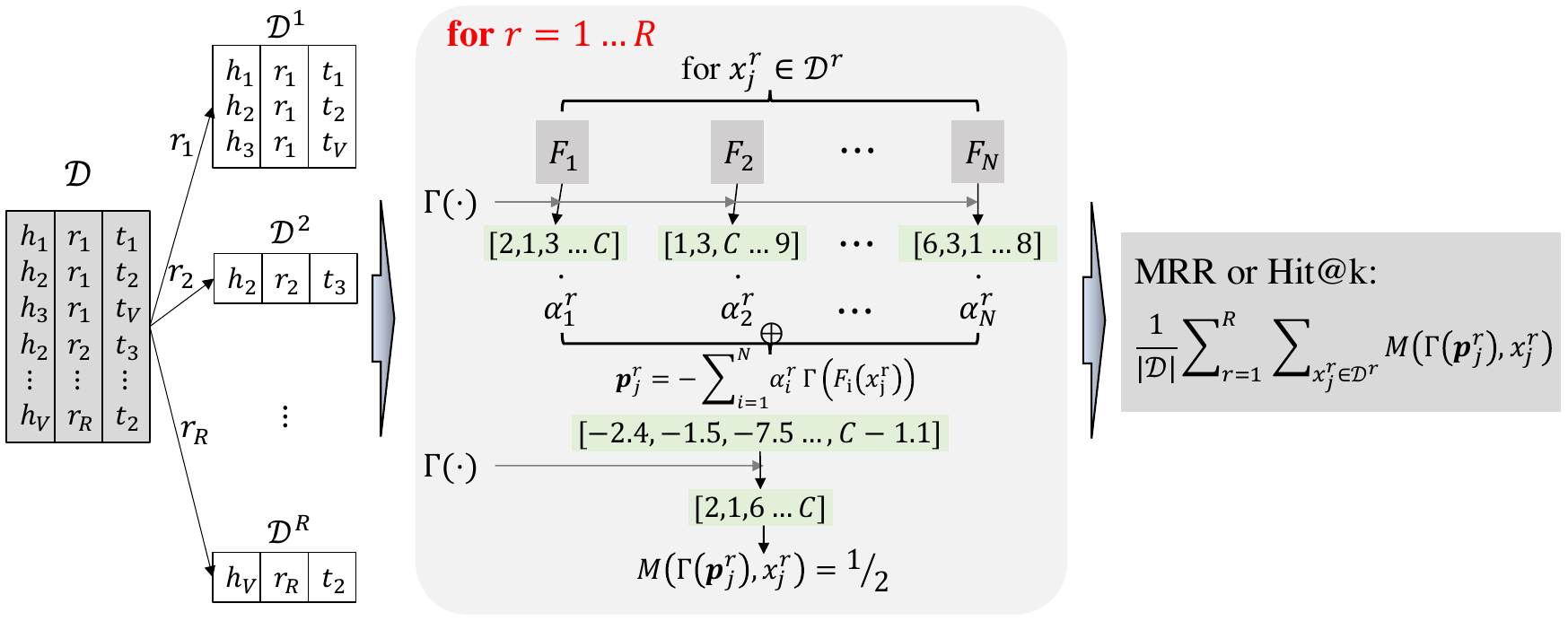}
    \caption{An overview of the relation-wise ensemble problem.}
    \label{fig:overview}
  \end{figure}

  An overview of the relation-wise ensemble problem is provided
  
  in Figure~\ref{fig:overview}.
  First,
  the dataset $\mathcal D$ is split into multiple sub-sets $\mathcal D^1, \mathcal D^2,\dots, \mathcal D^R$
  according to the relations.
  For each sample $x^r_j$ from $\mathcal D^r$,
  the models $F_1, F_2, \dots, F_N$ output the scores
  and the ranking function $\Gamma(\cdot)$ provides ranking lists for the $C$ entities
  according to their scores.
  The relation-wise ensemble weights $\alpha^r_1,\alpha^r_2,\dots,\alpha^r_N$
  re-weight the rank lists as the new scores $\bm p_j^r$ of $x^r_j$
  and re-rank the new scores to evaluate the perfomance.

  Two metrics are used in this paper: (i) Mean reciprocal ranking (MRR):
  \begin{gather*}
    \text{MRR} = \frac{1}{2\left | \mathcal{D} \right |}\sum_{(h,r,t)\in\mathcal{D}}\big(\frac{1}{p_h}+\frac{1}{p_t}\big);
  \end{gather*}
  and (ii) Hit@$k$: ratio of ranks no larger than $k$, i.e.,
  \begin{gather*}
    \text{Hit}@k = \frac{1}{2\left | \mathcal{D} \right |}\sum_{(h,r,t)\in\mathcal{D}}\big(\mathbb{I}(p_h \leq k)+\mathbb{I}(p_t \leq k)\big),
  \end{gather*}
  where $\mathbb{I}(a) = 1$ if a is true, otherwise 0,
  $p_h$ is the rank of head entity $h$ in the head-prediction sub-task (the same for $p_t$ and $t$).
  The larger the MRR
  or Hit@$k$, the better is the embedding.

  \subsection{Relation Properties}
  \label{app:relationproperty}
  
  In the main text,
  we claim that the different models work properly for different relations.
  In this part, we 
  summarize the types of relations that different models
  can handle in Table~\ref{tab:model_ability}.
  
  \begin{table}[htbp]
    \centering
    \caption{The pattern modeling  and inference  abilities of selected score functions.}
    \small
    \begin{tabular}{c|c|c|c|c|c}
      \toprule
      Model   & Symmetry     & Antisymmetry         & Inversion             & Composition & Hierarchy \\ \midrule
      TransE  & \XSolidBrush  &  \Checkmark         &   \Checkmark & \Checkmark      &   \XSolidBrush  \\
      RotatE  &  \Checkmark  & \Checkmark          & \Checkmark   & \Checkmark      & \XSolidBrush   \\
      HousE  &  \Checkmark  & \Checkmark          & \Checkmark   & \Checkmark      & \XSolidBrush   \\
      ComplEx & \Checkmark  & \Checkmark          &  \Checkmark   &  \XSolidBrush          &  \Checkmark     \\ 
      ConvE &  \Checkmark  & \Checkmark          &  \XSolidBrush   &  \XSolidBrush          &  \Checkmark  \\ 
      CompGCN &   \Checkmark  & \Checkmark          &  \Checkmark   &  \XSolidBrush          &  \XSolidBrush   \\
      \bottomrule
    \end{tabular}
    \label{tab:model_ability}
  \end{table}
  
  In addition, we show the performance of various base models for specific relations
  on WN18RR.
  
  Among the total of eleven relations considered, 
  we present the results based on the following four representative relations:
  \begin{itemize}
    \item 
     membe\_meronym: translational models such as TransE and RotatE exhibit the highest performance.
    \item
     synset\_domain\_topic\_of: bilinear models like ComplEx achieve the best results.
    \item 
     has\_part: while traditional scoring functions perform well on this relation, neural network-based models such as ConvE and CompGCN exhibit suboptimal performance.
    \item 
     verb\_group: in contrast to has\_part, neural network models such as ConvE and CompGCN perform better, whereas traditional scoring functions show inferior performance.
  \end{itemize}
  These results demonstrate that KG embedding models may specialize in different relations,
  leading to significant variation in their performance across relations.
  
  \begin{figure}[H]
    \centering
    \includegraphics[width=0.7\columnwidth]{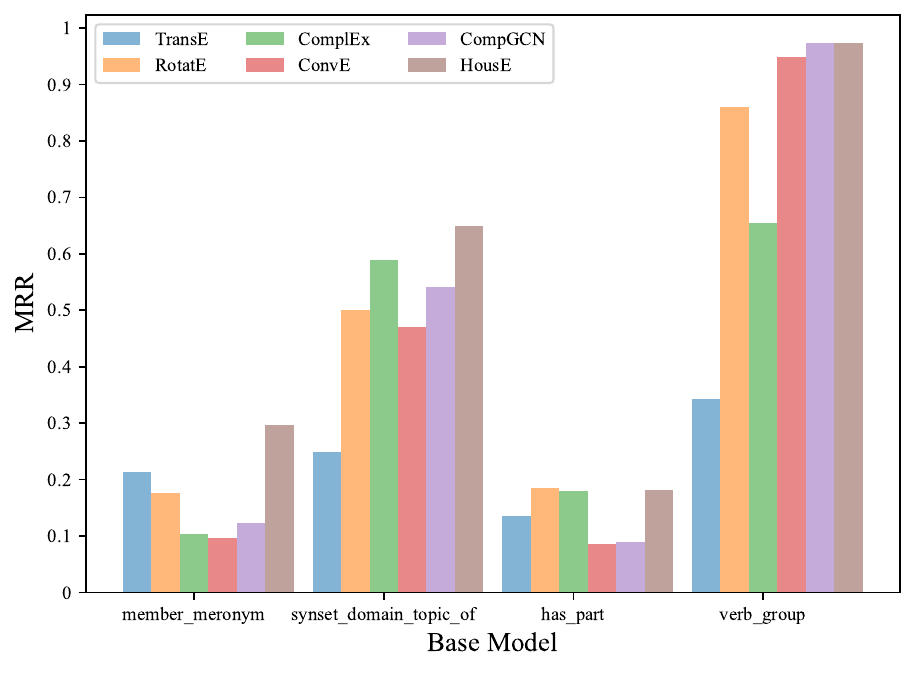}
    \caption{MRR of selected base models for specific relations on the WN18RR dataset.}
    \label{fig:relation_mrr}
  \end{figure}

  \section{Supplementary Materials for the Experimental Settings}
  
  \subsection{Statistics of Datasets}
  \label{app:datasets}

  We use the following datasets for evaluation:
  (i) WN18RR is a link prediction dataset which is a subset of WordNet~\cite{dettmers2018convolutional};
  (ii) FB15k-237 contains triplets of knowledge base relationships and textual mentions of Freebase entity pairs~\cite{toutanova2015observed};
  (iii) NELL-995 is a dataset built from the web via an intelligent agent called Never-Ending Language Learner
  that reads the web over time~\cite{xiong2017deeppath};
  (iv) ogbl-biokg is a KG, which was created using data from a large number of biomedical data repositories~\cite{hu2020open};
  and (v) ogbl-wikikg2 is a KG extracted from the Wikidata knowledge base~\cite{hu2020open}.
  Statistics of these datasets are provided in Table~\ref{tab:datainfo}.
  
  \begin{table}[htbp]
    \centering
    \caption{Statistics of the datasets.}
    \begin{tabular}{c|ccccc}
      \toprule
      Dataset      & $|\mathcal V|$ & $|\mathcal R|$ & $|\mathcal D_\text{tra}|$ & $|\mathcal D_\text{val}|$ & $|\mathcal D_\text{tst}|$ \\ \midrule
      WN18RR       & 40,943         & 11             & 86,835                    & 3,034                     & 3,134                     \\
      FB15k-237    & 14,541         & 237            & 272,115                   & 17,535                    & 20,466                    \\
      NELL-995     & 74,536         & 200            & 149,678                   & 543                       & 2,818                     \\ \hline
      ogbl-biokg   & 93,773         & 51             & 4,762,678                 & 162,886                   & 162,870                   \\
      ogbl-wikikg2 & 2,500,604      & 535            & 16,109,182                & 429,456                   & 598,543                   \\
      \bottomrule
    \end{tabular}
    \label{tab:datainfo}
  \end{table}

  \subsection{Hyperparameter Setting}
  \label{app:hp}
  
  We list the hyperparameters for base models in KGTuner~\cite{zhang2022kgtuner}\footnote{\url{https://github.com/LARS-research/KGTuner}} on the
  WN18RR, FB15k-237 and NELL-995 datasets in Table~\ref{tab:hp_wn18rr} and Table~\ref{tab:hp_fb15k237}.
  
  \begin{table*}[ht]
    \centering
    \caption{Hyperparameters for the WN18RR dataset.}
    \label{tab:hp_wn18rr}
    \small
    \begin{tabular}{c|c|c|c|c}
      \toprule
      HP/Model            & ComplEx               & ConvE                  & TransE                 & RotatE                 \\
      \midrule
      \# negative samples & 32                    & 512                    & 128                    & 2048                   \\
      \midrule
      loss function       & BCE\_mean             & BCE\_adv               & BCE\_adv               & BCE\_adv               \\
      gamma               & 2.29                  & 12.16                  & 3.50                   & 3.78                   \\
      adv. weight         & 0.00                  & 0.78                   & 1.14                   & 1.66                   \\
      \midrule
      regularizer         & NUC                   & DURA                   & FRO                    & FRO                    \\
      reg. weight         & $1.21 \times 10^{-3}$ & $ 9.79 \times 10^{-3}$ & $ 4.19 \times 10^{-4}$ & $5.13 \times 10^{-8}$  \\
      dropout rate        & 0.28                  & 0.02                   & 0.00                   & 0.00                   \\
      \midrule
      optimizer           & Adam                  & Adam                   & Adam                   & Adam                   \\
      learning rate       & $6.08 \times 10^{-4}$ & $ 6.88 \times 10^{-4}$ & $ 1.02 \times 10^{-4}$ & $ 1.24 \times 10^{-3}$ \\
      initializer         & x\_uni                & x\_uni                 & norm                   & norm                   \\
      \midrule
      batch size          & 1024                  & 512                    & 512                    & 512                    \\
      dimension size      & 2000                  & 1000                   & 1000                   & 1000                   \\
      inverse relation    & False                 & False                  & False                  & False                  \\
      \bottomrule
    \end{tabular}
  \end{table*}

  \begin{table*}[ht]
    \centering
    \caption{Hyperparameters for the FB15k-237  and NELL-996 datasets.}
    \label{tab:hp_fb15k237}
    \small
    \begin{tabular}{c|c|c|c|c}
      \toprule
      HP/Model            & ComplEx               & ConvE                 & TransE                & RotatE                \\ \midrule
      \# negative samples & 512                   & 512                   & 512                   & 128                   \\    \midrule
      loss function       & BCE\_adv              & BCE\_sum              & BCE\_adv              & BCE\_adv              \\
      gamma               & 13.05                 & 14.52                 & 6.76                  & 14.46                 \\
      adv. weight         & 1.93                  & 0.00                  & 1.99                  & 1.12                  \\
      \midrule
      regularizer         & DURA                  & DURA                  & FRO                   & NUC                   \\
      reg. weight         & $9.75 \times 10^{-3}$ & $6.42 \times 10^{-3}$ & $2.16 \times 10^{-4}$ & $2.99 \times 10^{-4}$ \\
      dropout rate        & 0.22                  & 0.07                  & 0.02                  & 0.01                  \\
      \midrule
      optimizer           & Adam                  & Adam                  & Adam                  & Adam                  \\
      learning rate       & $9.70 \times 10^{-4}$ & $2.09\times 10^{-4}$  & $ 2.66\times 10^{-4}$ & $5.89\times 10^{-4}$  \\
      initializer         & uni                   & norm                  & x\_norm               & norm                  \\
      \midrule
      batch size          & 1024                  & 1024                  & 512                   & 1024                  \\
      dimension size      & 2000                  & 500                   & 1000                  & 2000                  \\
      inverse relation    & False                 & False                 & False                 & False                 \\
      \bottomrule
    \end{tabular}
  \end{table*}

  For CompGCN~\cite{vashishth2019composition}\footnote{\url{https://github.com/malllabiisc/CompGCN}}, 
  we use 200-dimensional embeddings for node and relation embeddings
  and apply the standard binary cross entropy loss with label smoothing.
  The number of GCN layers is 2,
  and the score function used in CompGCN is ConvE,
  the learning rate is set to 0.001,
  the batch size is 128, and the dropout rate is 0.1.
  
For HousE~\cite{li2022house}\footnote{\url{https://github.com/rui9812/HousE}}, 
we used the default hyperparameters specified in the original paper. 
Both node and relation embeddings were set to 800 dimensions. 
The learning rate was set to 0.0005, and the batch size was 1000.

  For the top three methods on the OGB leaderboard
  \footnote{\url{https://ogb.stanford.edu/docs/leader_linkprop/\#ogbl-biokg} and \url{https://ogb.stanford.edu/docs/leader_linkprop/\#ogbl-wikikg2}}, 
  since their code has been officially made public by OGB, 
  we used their code directly with their corresponding hyperparameters.
  

  \subsection{Details of Tree-structured Parzen Estimator (TPE)}
  \label{app:tpe}
  
  The TPE (Tree-structured Parzen Estimator) algorithm is a Bayesian optimization~\cite{snoek2012practical} method that aims to efficiently optimize black-box functions with a limited budget of function evaluations. It was introduced by~\cite{bergstra2011algorithms} as a part of the Hyperopt framework~\cite{bergstra2015hyperopt}, which focuses on hyperparameter optimization.
  Bayesian optimization is a sequential model-based optimization technique that leverages prior knowledge and data to intelligently search for the optimal solution. It uses a probabilistic surrogate model, typically a Gaussian process or a tree-based model, to model the unknown objective function. This model is iteratively updated as new observations are made, providing an estimate of the function's behavior and uncertainty.
  
  The TPE algorithm improves upon traditional Bayesian optimization by employing a novel method for modeling and sampling from the posterior distribution of the objective function. 
  It uses a tree-structured Parzen estimator to model the distribution of good and bad parameter configurations. 
  The algorithm maintains two density functions: 
  $p\left(x|y\right)=\ell(x) {\text{ if } y < y^{\ast }}$ otherwise $p\left(x|y\right)=g(x)$,
  where $\ell(x)$ is the density formed by using observations $\{x^{(i)}\}$ from past evaluations, 
  such that the corresponding loss (i.e., the performance metric for the model) is less than some threshold $y^{\ast}$ that lead to good results, 
  and $g(x)$ is the density formed by using the remaining observations that lead to bad results. 
  
  At each iteration, the TPE algorithm samples promising configurations from the good density and less promising configurations from the bad density. The algorithm balances the exploration-exploitation trade-off by dividing the sampled configurations into two groups based on their relative performance. The better-performing configurations are used to update the density function for good configurations, while the less successful ones update the density function for bad configurations. This process aims to guide the search towards promising regions of the parameter space.
  By iteratively updating the density functions and adaptively sampling configurations, the TPE algorithm efficiently explores the parameter space, gradually converging towards the optimal solution. It has been shown to be effective in hyperparameter optimization for machine learning models and other optimization tasks.
  
  Overall, the TPE algorithm, as a Bayesian optimization method, offers a principled and efficient approach for optimizing black-box functions with limited resources, making it particularly useful in scenarios where function evaluations are time-consuming or costly.
  
  \section{Supplementary Materials for Experimental Results}
  
%

  \subsection{Full Results of Learning Curves}
  \label{app:efficiency}
  
\begin{figure}[h]
    \centering
    \vspace{-7.5pt}
    
    \includegraphics[width=0.92\columnwidth]{images/learning_curve_legend.pdf}
    
    \subfigure[WN18RR with  $Q= 50$]
    {\includegraphics[width=0.37\columnwidth]{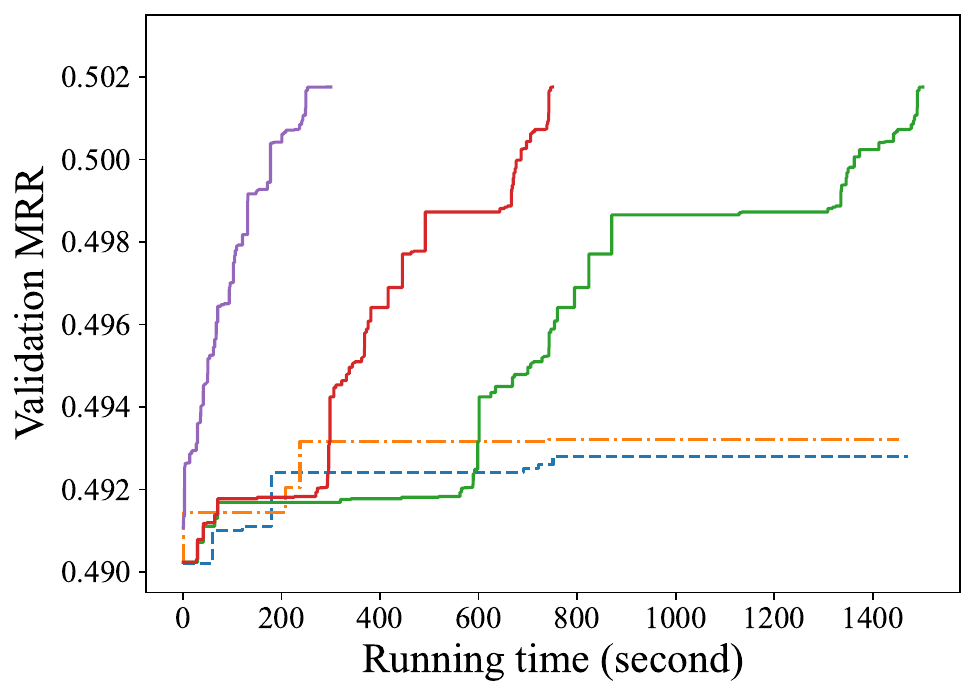}}
    \quad
    \subfigure[WN18RR with $Q= 100$]
    {\includegraphics[width=0.37\columnwidth]{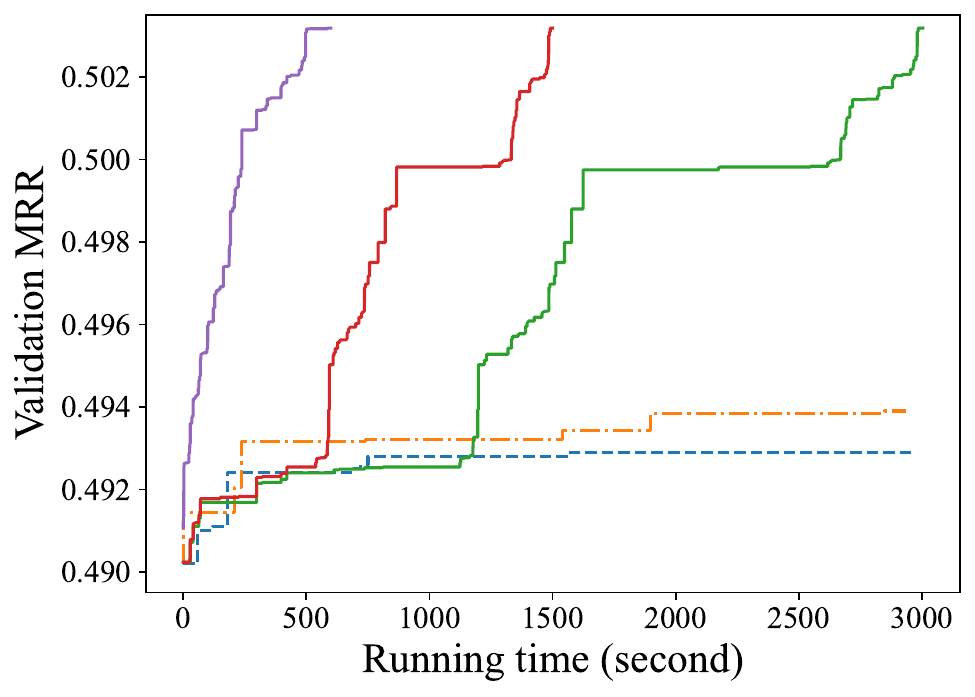}}
    
    \vspace{-7.5pt}
    \subfigure[FB15k-237 with  $Q= 50$]
    {\includegraphics[width=0.37\columnwidth]{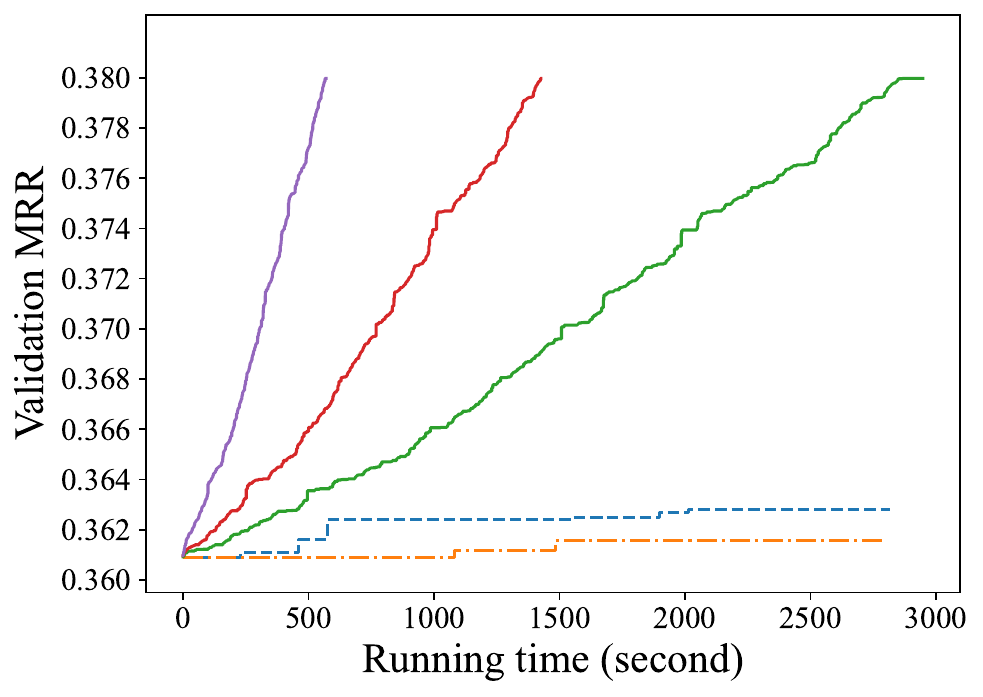}}
    \quad
    \subfigure[FB15k-237 with $Q= 100$]
    {\includegraphics[width=0.37\columnwidth]{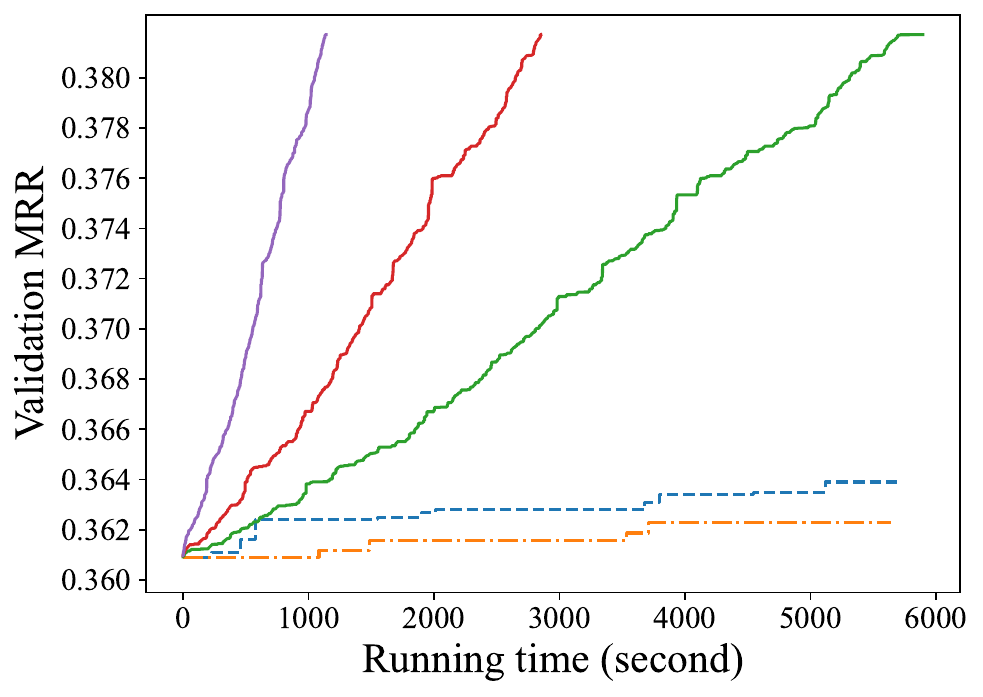}}
    
    \vspace{-7.5pt}	
    \subfigure[NELL-995 with  $Q= 50$]
    {\includegraphics[width=0.37\columnwidth]{images/learning_curve_nell_50.pdf}}
    \quad
    \subfigure[NELL-995 with $Q= 100$]
    {\includegraphics[width=0.37\columnwidth]{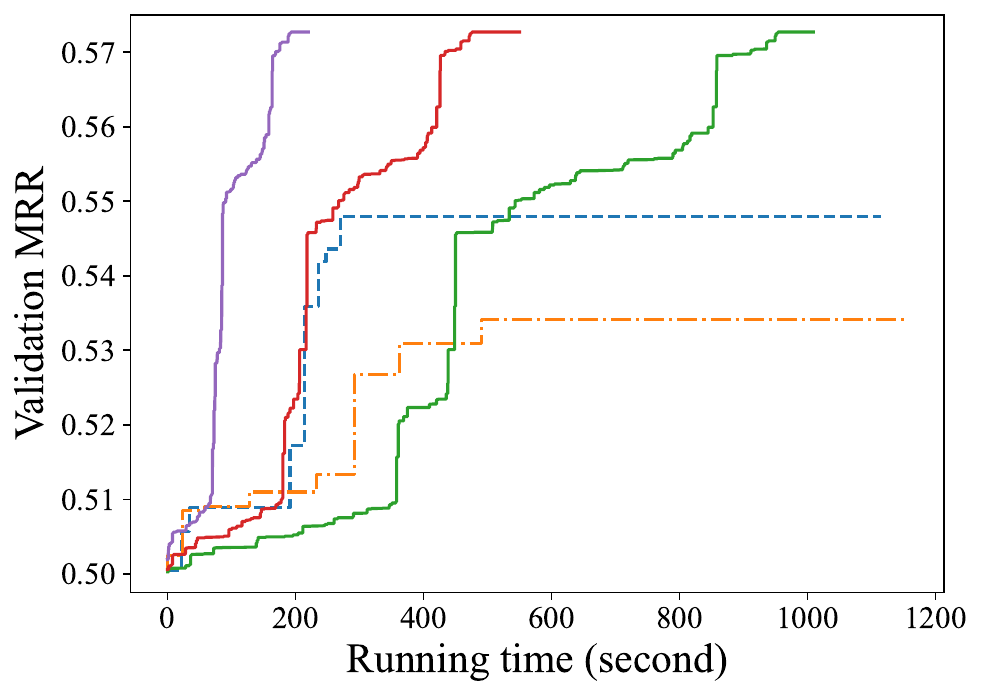}}
    
    \vspace{-7.5pt}	
    \subfigure[ogbl-wikikg2 with  $Q= 50$]
    {\includegraphics[width=0.37\columnwidth]{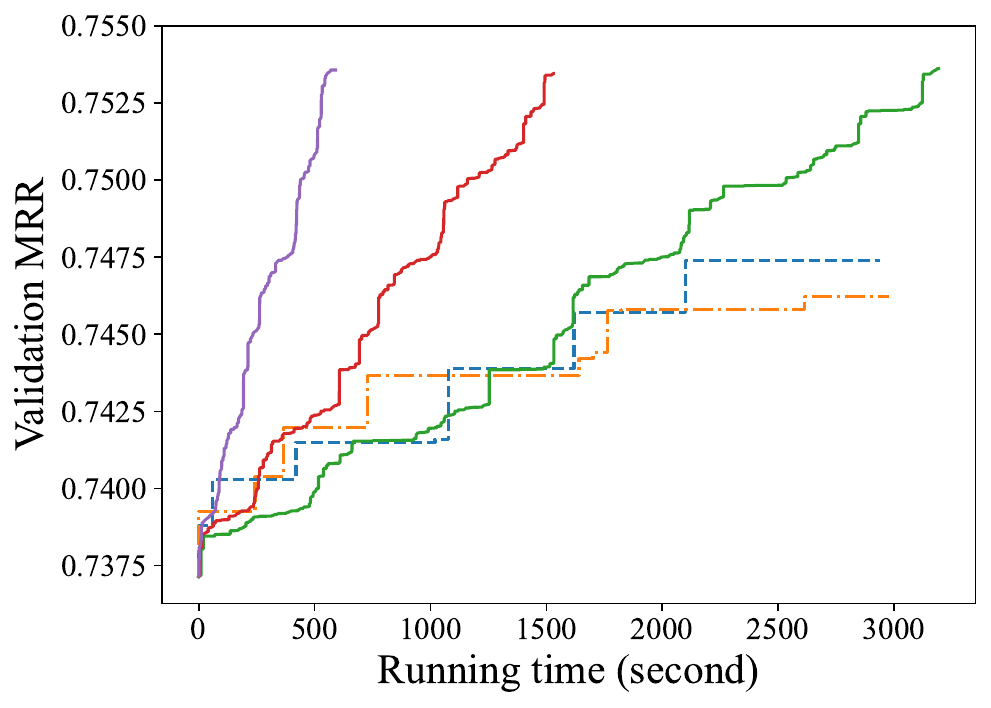}}
    \quad
    \subfigure[ogbl-wikikg2 with $Q= 100$]
    {\includegraphics[width=0.37\columnwidth]{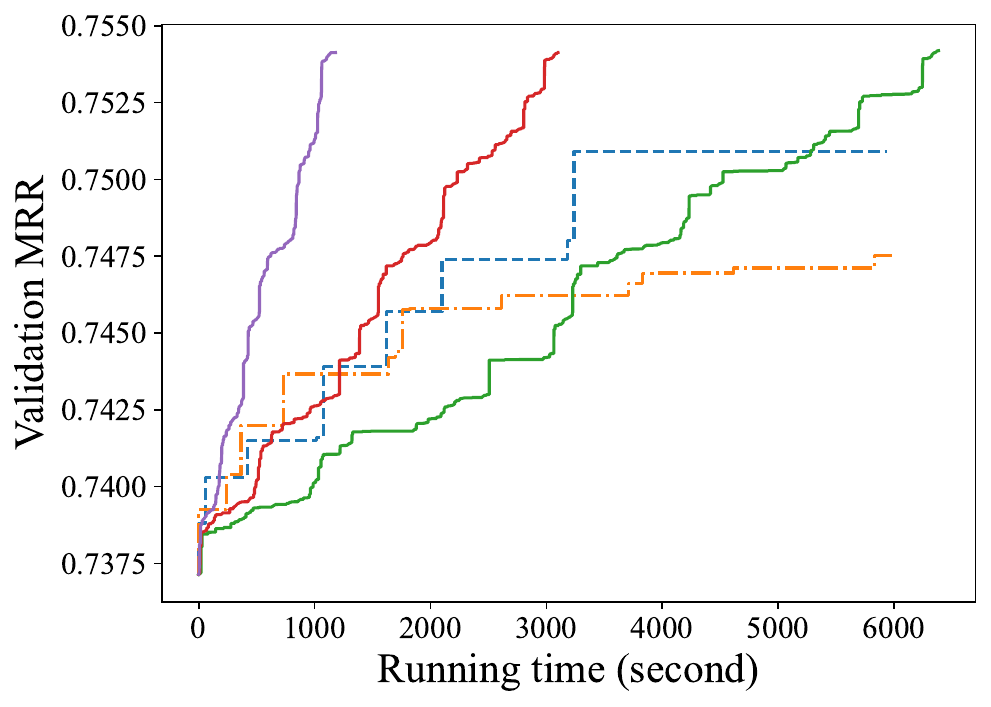}}
    
    \vspace{-7.5pt}	
    \subfigure[ogbl-biokg with  $Q= 50$]
    {\includegraphics[width=0.37\columnwidth]{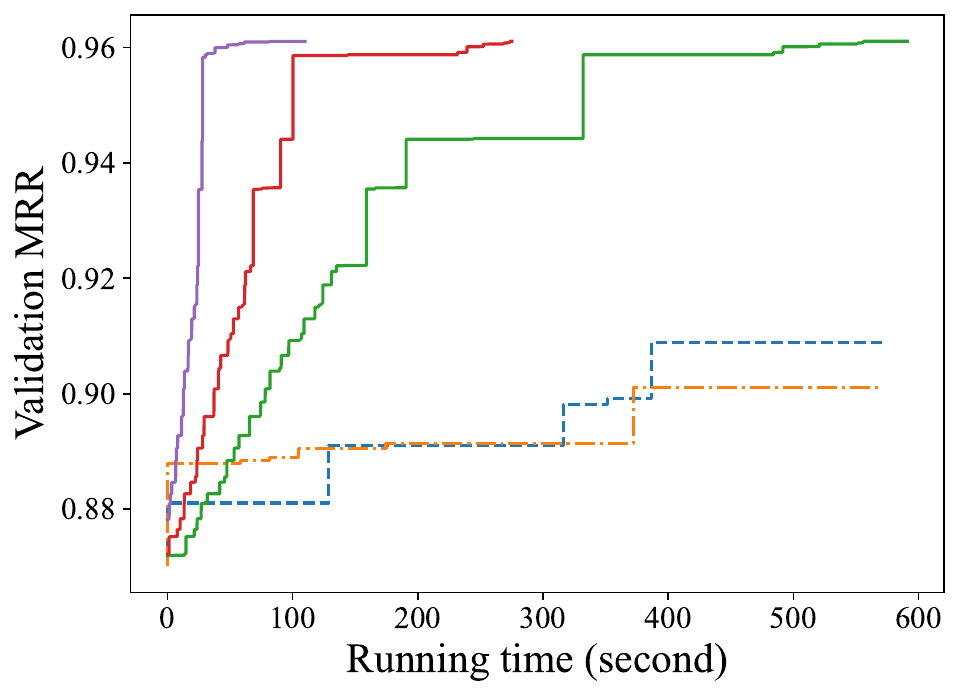}}
    \quad
    \subfigure[ogbl-biokg with $Q= 100$]
    {\includegraphics[width=0.37\columnwidth]{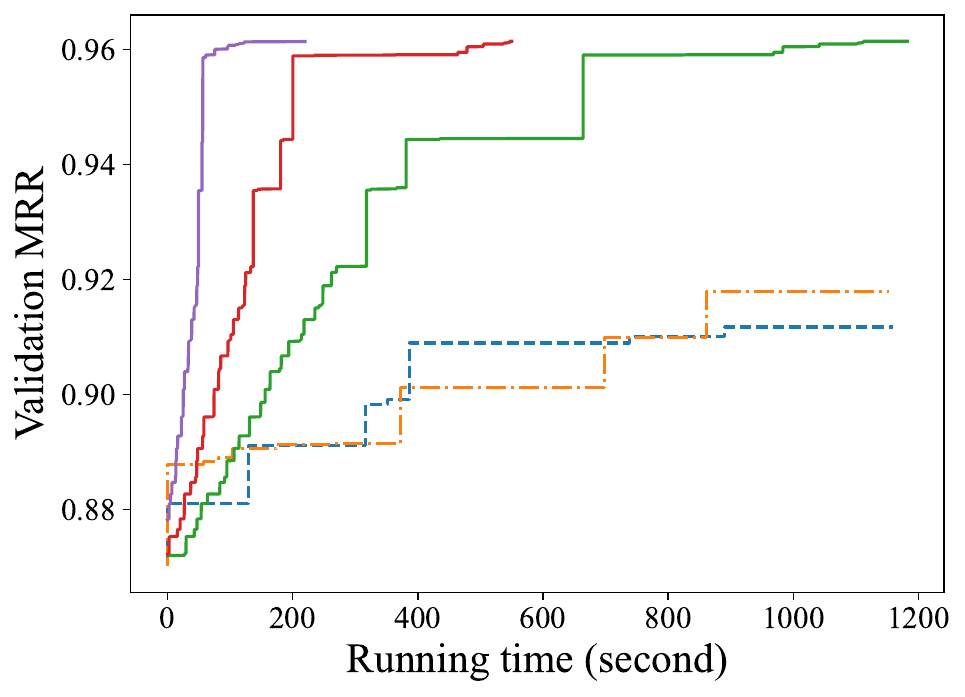}}

    \vspace{-10pt}
    \caption{Learning curves of different ensemble methods.
       \textit{RelEns-DSC2/5} indicates the number of threads used for parallel computing.}
    \vspace{-7.5pt}
    \label{fig:all_learning_curve}
  \end{figure}

  We provide the learning curves on WN18RR, FB15k-237 and NELL-995 datasets
  in this part.
  Figure~\ref{fig:all_learning_curve} confirms that the observations in Section~\ref{sec:efficiency} are consistent with the results presented here. While  \textit{RelEns-Basic} exhibits a superior modeling ability compared to  \textit{SimpleEns} by learning relation-specific weights, enabling different models to specialize in different relations. 
  However, the search complexity of  \textit{RelEns-Basic} increases exponentially with the number of relations. 
  \begin{itemize}
    \item On WN18RR with only 11 relations, \textit{RelEns-Basic} can slightly outperform \textit{SimpleEns}
    since the complexity is not increased much 
    and the relation-wise ensemble problem can have better performance than the general ensemble problem.
    \item Nevertheless, 
    on datasets containing hundreds of relations such as FB15k-237 and NELL-995 
    \textit{RelEns-Basic} exhibits significantly poorer performance compared to  \textit{SimpleEns} due to high sampling complexity in the relation-wise ensemble problem.
  \end{itemize}
  
  \clearpage

  \subsection{Ablation Study on Ensemble methods}
  \label{app:ablation}
  
  All the variants use the same way to ensemble predictions of base models.
  The same as mentioned in Section~\ref{ssec:relation-wise},
  we use the ensemble weights on rankings of base models 
  as the ensemble score,
  i.e., $\bm p_j^r =
  - \sum\nolimits_{i = 1}^N \alpha_i^r\Gamma\big(F_i( x_j^r )\big)$.
  The implementation details of the ensemble variants are provided as follows:
  \begin{itemize}
    \item Mean:	
      This is the most basic ensemble method which directly takes the arithmetic mean of the predictions made by each individual model.
      
    \item Stacking~\cite{wolpert1992stacked}:
    The stacking ensemble variant is a more sophisticated approach 
    that involves training a meta-model to learn how to combine the predictions of multiple base models. 
    Specifically, the rank lists outputted by the base models are used as features to train the meta-model. 
    Here, we use a logistic regression model implemented by scikit-learn~\cite{pedregosa2011scikit} as the meta-model, 
    with a maximum iteration of 300.
    Since the ranking metrics on the validation data is non-differentiable,
    we use the same loss function during training on the validation data
    to optimize the parameters of meta-model.
  
    \item SimpleEns: 
    This variant is a degenerated problem of \eqref{eq:no-split}
    by setting identical ensemble weights of different relations on a single mode,
    i.e., $\alpha_i^1=\alpha_i^2=\cdots = \alpha_i^{R}$ for $i=1,\dots,N$.
    In order to keep consistent with RelEns-Basic and RelEns-DSC,
    these weights are searched by optimizing \eqref{eq:no-split} with TPE technique.
    
    \item MRR-Mean:
    The MRR-Mean variant incorporates the Mean Reciprocal Rank (MRR) of the base model as a weighting factor. 
    In contrast to the Mean variant, it assigns proportionally greater weight to the superior individual base model.
  
  \end{itemize}
  
  We conducted an ablation study on the WN18RR, FB15k-237, and NELL-995 datasets, and the results are shown in Table~\ref{tab:ensemble2}.
  MRR can provide a general indication of the importance of different models, 
  but higher MRR does not always correlate with higher model importance due to the crucial role of diversity among the base models in the ensemble strategy. 
  Additionally, MRR-weight may not be optimal weights, 
  necessitating further weight searches to enhance performance.
  The results consistently demonstrate the superiority of searched methods (RelEns-DSC) over MRR-weighted methods (MRR-Mean).
  
  \begin{table*}[ht]
      \centering
      \caption{Comparison of ensemble methods on WN18RR, FB15k-237 and NELL-995 datasets.}
      \small
      \setlength\tabcolsep{3.8pt}
      \begin{tabular}{c|cccc|cccc|cccc}
          \toprule
          Dataset      & \multicolumn{4}{c|}{WN18RR} & \multicolumn{4}{c|}{FB15k-237} & \multicolumn{4}{c}{NELL-995}                                                                                                                                                                   \\ \midrule
          Model        & MRR                         & Hit@1                          & Hit@3                        & Hit@10          & MRR             & Hit@1           & Hit@3           & Hit@10          & MRR             & Hit@1           & Hit@3           & Hit@10          \\ \midrule
          Mean         & 0.4963                    & 0.4531                       & 0.5175                       & 0.5746          & 0.3572           & 0.2643          & 0.3918          & 0.5457          & 0.5412           & 0.4661           & 0.5823          & 0.6622          \\
          Stacking     & 0.4952                    & 0.4518                      & 0.5169                       & 0.5751          & 0.3563          & 0.2639          & 0.3921          & 0.5433          & 0.5365          & 0.4637          & 0.5851          & 0.6669          \\
          SimpleEns    & 0.5121                & 0.4670                        & 0.5289                     & 0.6021          & 0.3621          & 0.2683          & 0.3977          & 0.5525          & 0.5416          & 0.4758          & 0.5823          & 0.6601          \\
          MRR-Mean    & 0.5143                & 0.4697                        & 0.5311                     & 0.6028          & 0.3645          & 0.2702          & 0.3993          & 0.5547          & 0.5460          & 0.4732          & 0.5834          & 0.6604          \\
          RelEns-DSC & \textbf{0.5201}   & \textbf{0.4770}  & \textbf{0.5375}  & \textbf{0.6039} & \textbf{0.3680} & \textbf{0.2746} & \textbf{0.4046} & \textbf{0.5554} & \textbf{0.5499} & \textbf{0.4823} & \textbf{0.5901} & \textbf{0.6609} \\ \bottomrule
      \end{tabular}
      \label{tab:ensemble2}
  \end{table*}
  
  \end{document}